\pdfoutput=1
\documentclass[11pt]{article}

\usepackage[]{acl}

\usepackage{times}
\usepackage{latexsym}
\usepackage{adjustbox}
\usepackage[linesnumbered,ruled,vlined]{algorithm2e}
\usepackage{float}
\usepackage{color}
\usepackage{colortbl}
\definecolor{hiscolor}{RGB}{220,224,228}
\usepackage{multirow}
\usepackage{enumitem}
\usepackage{xcolor}
\usepackage{booktabs}
\usepackage{array}
\usepackage{makecell}
\usepackage{amsmath}
\usepackage{amsfonts}       % blackboard math symbols
\usepackage{tcolorbox}
\tcbuselibrary{listings, breakable, skins, theorems}
\renewcommand{\labelitemii}{+}
\newcounter{prompt}

\newtcolorbox[auto counter, number within=section]{promptbox}[2][]{
colback=gray!10!white,
colframe=gray!60!gray,
fonttitle=\bfseries\sffamily,
title=Prompt~\thetcbcounter: #2,
rounded corners,
arc=1.3mm,
boxrule=0.5pt,
enhanced,
breakable,
listing only,
listing options={
    basicstyle=\ttfamily\bfseries\itshape\fontsize{5}{6},
    numbers=left,
    numberstyle=\tiny\color{gray!80!black},
    stepnumber=1,
    numbersep=5pt,
    showspaces=false,
    showstringspaces=false
},
label={prompt:#1}
}

\usepackage[T1]{fontenc}
\usepackage[utf8]{inputenc}

\usepackage{microtype}

\usepackage{inconsolata}

\usepackage{graphicx}

\title{LLaMA-E: Empowering E-commerce Authoring with Object-Interleaved Instruction Following}

\author{Kaize Shi\textsuperscript{\rm 1},
        Xueyao Sun\textsuperscript{\rm 1,\rm 2}, 
        Dingxian Wang\textsuperscript{\rm 1}, 
        Yinlin Fu\textsuperscript{\rm 3}, 
        Guandong Xu\textsuperscript{\rm 1}\thanks{~~Corresponding author},
        Qing Li\textsuperscript{\rm 2} \\
        \textsuperscript{\rm 1}University of Technology Sydney \\ 
        \textsuperscript{\rm 2}The Hong Kong Polytechnic University\\
        \textsuperscript{\rm 3}Etsy\\
        \texttt{\{Kaize.Shi, Guandong.Xu\}@uts.edu.au}}

\begin{document}
\maketitle
\begin{abstract}
E-commerce authoring entails creating engaging, diverse, and targeted content to enhance preference elicitation and retrieval experience. While Large Language Models (LLMs) have revolutionized content generation, they often fall short in e-commerce applications due to their limited memorization of domain-specific features. This paper proposes LLaMA-E, the unified e-commerce authoring models that address the contextual preferences of customers, sellers, and platforms, the essential objects in e-commerce operation. We design the instruction set derived from tasks of ads generation, query-enhanced product title rewriting, product classification, purchase intent speculation, and general e-commerce Q\&A. The instruction formulation ensures the interleaved cover of the presented and required object features, allowing the alignment of base models to parameterise e-commerce knowledge comprehensively. The proposed LLaMA-E models achieve state-of-the-art evaluation performance and exhibit the advantage in zero-shot practical applications. To our knowledge, this is the first LLM tailored to empower authoring applications with comprehensive scenario understanding by integrating features focused on participated objects.~\footnote{The LLaMA-E is released at \url{https://huggingface.co/DSMI/LLaMA-E\#/}, with the demo available at \url{https://huggingface.co/spaces/KaizeShi/LLaMA-E\#/}.}
\end{abstract}

% ~\footnote{The LLaMA-E is released at \url{https://huggingface.co/DSMI/LLaMA-E\#/}, with the demo available at \url{https://huggingface.co/spaces/KaizeShi/LLaMA-E\#/}.}

% ~\footnote{The LLaMA-E is released at \url{https://anonymous.4open.science/r/LLaMA-E-C392/}. The demo is online, and we will release the address after acceptance.}

\section{Introduction}

E-commerce authoring encompasses creating diverse and innovative textual content for online services, such as product copywriting, advertisements, and Q\&A~\cite{zhang2022product}. Automatically generating authoring content can enhance the product retrieval experience, improve preference elicitation, and drive sales and conversions~\cite{Jing2023Stylized}. Present task-specific authoring models predominantly focus on independent features, missing the capacity to interleave the features of objects in interactive e-commerce scenarios. These limitations constrain the model's understanding of e-commerce operations, disregarding their potential to fit and apply positively promoted features in the fine-grained authoring tasks~\cite{chan-etal-2020-selection}.

\begin{figure}[tbp]
 \centering
 \includegraphics[width=0.5\textwidth]{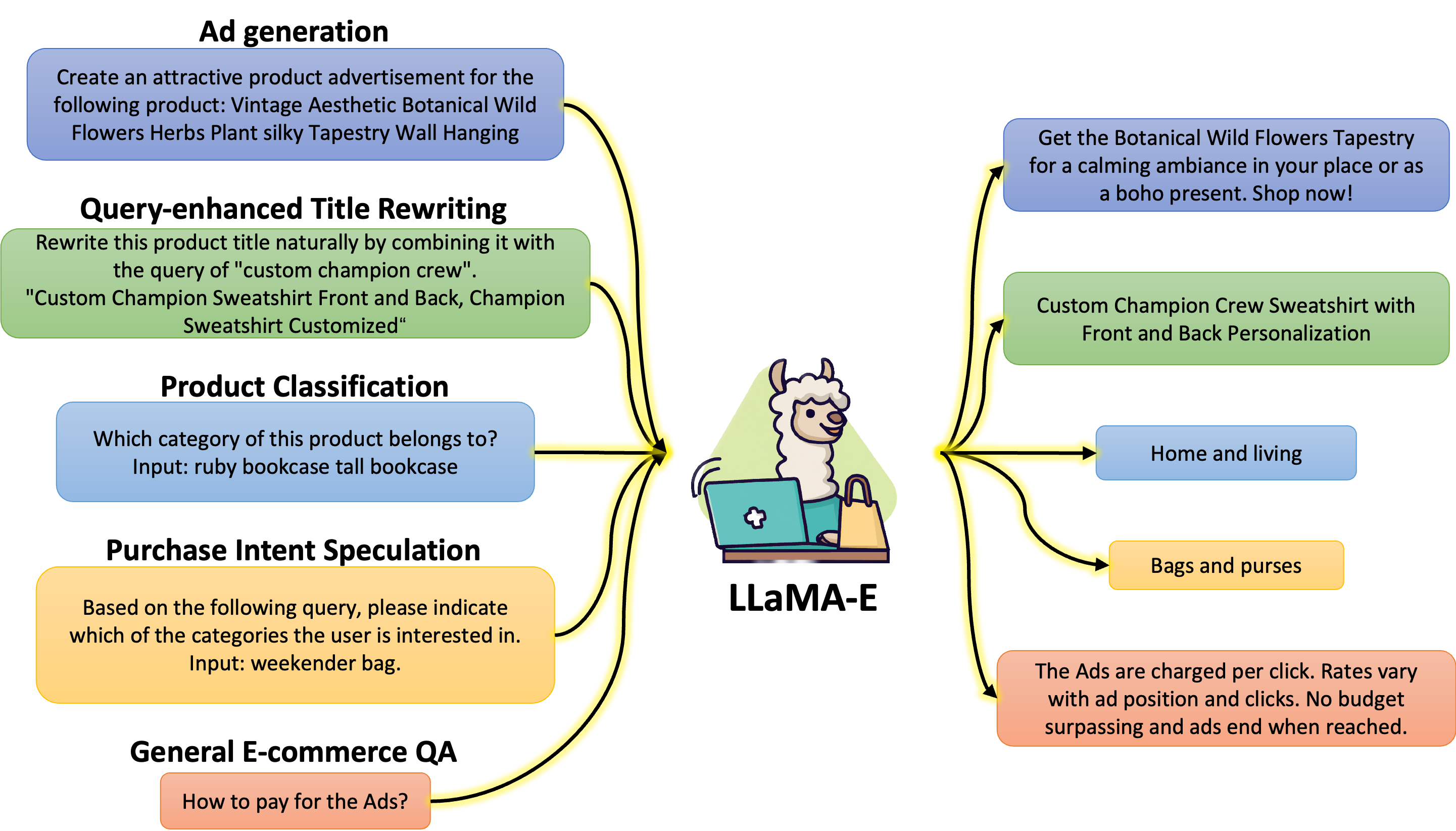}
 \caption{We train the LLaMA-E models based on the instructions set of various e-commerce authoring tasks, which interleaved integrating the object features for enhancing the comprehensive scenario understanding\protect\footnotemark}
 \label{fig:AMR_example}
\end{figure}
\footnotetext{The LLaMA-E icon was generated by DALL$\cdot$E: \url{https://labs.openai.com/}.}

Natural language processing (NLP) has witnessed a significant transformation with the emergence of the instruction-following large language models (LLMs)~\cite{zhao2023survey}. These powerful models have revolutionized how NLP tasks are approached, introducing a unified paradigm with potential for advancements~\cite{mialon2023augmented}. LLMs, such as the widely known ChatGPT~\footnote{\url{https://www.openai.com}}, acquire a broad spectrum of knowledge trained on vast corpora, enabling them to demonstrate remarkable generation performance and deliver impressive results in numerous applications, such as information retrieval, controlled generation, etc~\cite{bao2023large, shi-etal-2023-amr, shi2024compressing}. The comprehensive corpora allow LLMs to capture the logic of language representation and acquire a macro understanding of common sense and semantics. However, the general LLMs encounter challenges in comprehending and producing the intricacies of personalized and specialized scenarios due to data barriers that isolate long-tail domain-specific knowledge~\cite{zhao2023domain, pmlr-v202-kandpal23a}. Furthermore, certain LLMs rely on remote centralized services, which raises concerns regarding privacy protection in data transmission.

Comprehensively understanding complex e-commerce scenarios following the instructions integrating object-interleaved features offers significant opportunities to align LLMs in handling diverse authoring applications in a unified manner~\cite{lester2021power}. This procedure enables the general LLMs with common sense knowledge to focus on the e-commerce knowledge. Consequently, LLMs enhance the capacity for generalization and feature-fitting through contextually sensitive instructions, thereby releasing their ability for fine-grained downstream applications~\cite{singhal2023large}. Moreover, the customization of LLMs locally maximizes privacy by mitigating potential breaches related to sharing sensitive information during inference processes~\cite{Peris2023Privacy}.

This paper proposes the LLaMA-E, the instruction following LLMs specifically tailored for e-commerce authoring scenarios. Recent studies have shown that automatic self-instructional tuning can enhance the performance of LLMs in domain-specific applications by allowing them to generate content that closely follows the instructions and precisely meets the contextual expectations of the given scenario~\cite{wang-etal-2023-self-instruct, singhal2023large, thirunavukarasu2023large}. Inspired by this, we align LLMs to gain a thorough understanding of e-commerce authoring scenarios by injecting the knowledge featured by vital objects: sellers, customers, and platforms, avoiding feature bias arising from task-isolated learning. Specifically, domain experts are engaged to formulate the seed set to interleaved integrate object features, focusing on the tasks of ads generation, query-enhanced product title rewriting, product classification\footnote{The product taxonomy in this paper is defined as clothing, accessories, home and living, weddings, art and collectibles, craft supplies and tools, jewelry, paper and party supplies, toys and games, electronics and accessories, books movies and music, bath and beauty, bags and purses, shoes, pet supplies.}, query intent speculation, and general e-commerce Q\&A. After the raw instructions are collected, the teacher model, GPT-3.5-turbo-301, is introduced to expand the expert-defined task-specific instructions for enhancing the generalizability of model training. The seed instruction set is then combined with the expanded instruction set to the final instruction data, which consists of 120k instruction pairs after pruning. The LLaMA-E models are trained following the final instruction set and evaluated by the evaluation system designed from practical requirements to assess their effectiveness in empowering e-commerce authoring content presentation. The results demonstrate that LLaMA-E models achieve state-of-the-art performance, also surpassing general LLMs in held-out unseen tasks, proving their serviceability in real-world applications. The contributions of this paper are summarized as follows:
\begin{itemize}
\item We propose LLaMA-E, the LLMs designed specifically for uniformly presenting practical, object-oriented e-commerce authoring content to cater to various scenario objects.
\item We propose the e-commerce authoring instruction set that integrates object-interleaved features to prompt the alignment of LLMs to enable comprehensive scenario understanding.
\item The LLaMA-E models achieve state-of-the-art results compared with baselines. To the best of our knowledge, this is the first work in introducing LLMs to e-commerce authoring.
\end{itemize}

\section{Related Works}

\subsection{E-commerce Authoring}

E-commerce authoring aims to create diverse and engaging content to highlight product features and encourage purchases~\cite{guo2022automatic}. One straightforward approach is modifying the fixed patterns. Wang et al.~\cite{wang-etal-2017-statistical} proposed a statistical framework that generates product descriptions using templates extracted from product attributes. Xiao et al.~\cite{xiao2019text} generated summaries of product titles by defining the keyword categories. With advancements in NLG paradigms like Transformers~\cite{NIPS2017_3f5ee243}, models have improved in representing complex features and incorporating domain-specific details.

Recent research has focused on practical applications in e-commerce. Zhang et al.~\cite{zhang2022product} developed APCG, a system that uses human feedback to refine transformer-generated content, significantly improving click-through and conversion rates at JD.com. Wang et al.~\cite{wang-etal-2022-interactive} proposed generating descriptions by combining product titles, attributes, and marketer-created descriptions. Chen et al.~\cite{chen2019towards} integrated product aspects, user categories, and a knowledge base for personalized descriptions. In advertising, Chan et al.~\cite{chan-etal-2020-selection} generated ads by selecting representative products for the post topic, while Zhang et al.~\cite{zhang2022scenario} created a model for generating ads based on multiple products and scenario requirements.

\subsection{E-commerce Language Models}

E-commerce language models address various tasks to boost sales, user interaction, and personalized services~\cite{chen2023knowledge}. These tasks include auto Q\&A, product summarization, and sentiment analysis~\cite{Varia2023}. For instance, Zhang et al.~\cite{zhang2020bert} proposed E-BERT, a model incorporating phrase-level and product-level knowledge, improving Q\&A and product classification performance. Xu et al.~\cite{xu-etal-2021-k-plug} introduced K-PLUG, a pre-trained language model for generative tasks using product and e-commerce knowledge. Li et al.~\cite{li2024ecomgpt} developed EcomGPT, instructional fine-tuned BLOOMZ models that showed more competitive performance than ChatGPT on general e-commerce tasks.

Studies have also applied language models to enhance customized e-commerce services, such as recommender systems and information retrieval~\cite{liu2023pre}. Geng et al.~\cite{geng2022path} created a path language model for generating explainable product recommendations. Lu et al.~\cite{lu-etal-2021-graph} developed a multilingual retrieval model based on BERT to improve e-commerce search engines. Huang et al.~\cite{Huang2023} fine-tuned large language models on Amazon data to predict query similarity, which improves search ranking and matching accuracy.

\section{Methods}

\begin{figure*}[!htbp]
 \centering
 \includegraphics[width=1\textwidth]{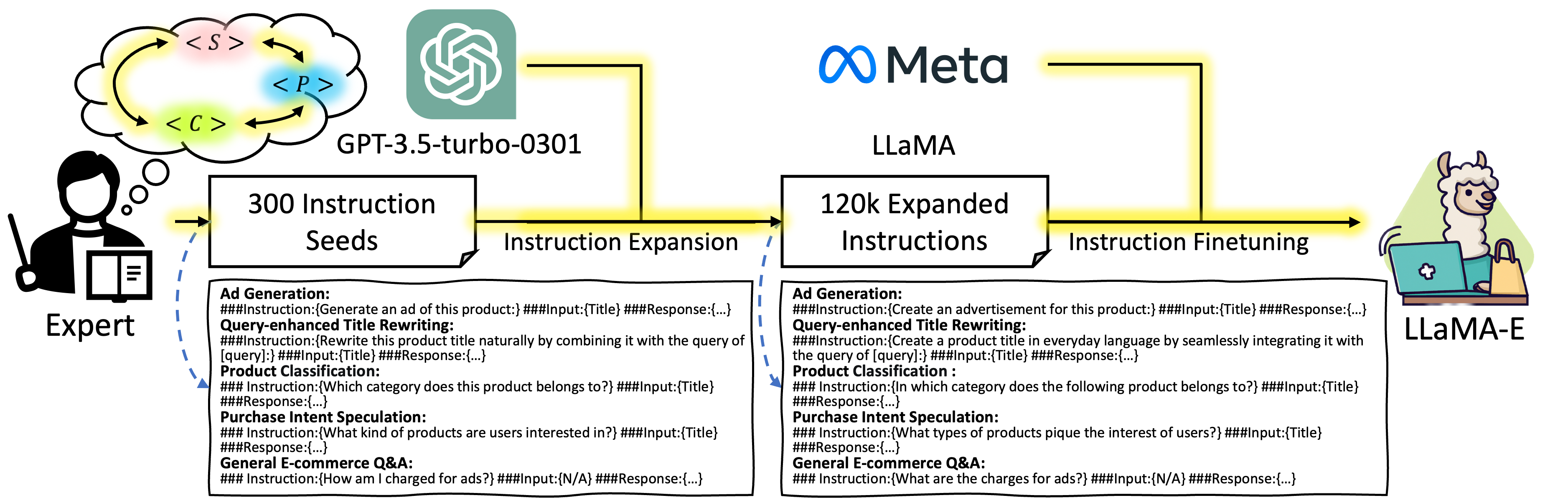}
 \caption{The development process of the LLaMA-E models, which includes the instruction formulating, instruction expansion, and instruction tuning for e-commerce authoring scenarios}
 \label{fig:method_overview}
\end{figure*}

The development process of the LLaMA-E models is illustrated in Figure~\ref{fig:method_overview}, including instruction formulating, expansion, and tuning. The following sections elaborate on each sub-process in detail.

\subsection{Instruction Formulating}

Formulating the informative instruction format requires integrating object-interleaved features from crucial e-commerce participants. This integration aims to align LLMs with comprehensive scenario understanding for executing authoring tasks. This paper focuses on the features of the seller, customer, and platform, which are the essential creators and consumers of e-commerce authoring content.

\textbf{Seller $<S>$}: The seller object significantly contributes to e-commerce authoring services by crafting attractive and informational product titles that encompass vital features, such as the product's name, style, brand, or model. These product titles provide an intuitive information channel for potential customers and effectively convey specific product features to the official platform.

\textbf{Customer $<C>$}: The customer object serves as the primary audience for e-commerce authoring services. They actively participate in the authoring process by providing personalized product preferences. The customer query corresponding to specific products is the vital textual carrier for associating the features of products and personalized preferences, which can be subdivided as follows:

\labelitemii \texttt{Explicit feature $<C_0>$}: This feature is intuitively reflected through the textual information in the customer query, which can provide specific feedback on the explicit features of the customer's intended product. The query text acts as an indicator of the specific features or attributes that the customer is retrieving in the product.

\labelitemii \texttt{Implicit feature $<C_1>$}: This feature encapsulates the potential purchase interest that can be inferred from the user query, thereby supporting the authoring process correlated with the specific customer intention. The features are semantically abstracted from the query text and can be elicited to associate with other features from different objects based on specific authoring scenarios.

\textbf{Platform $<P>$}: As the service provider of the e-commerce authoring models, features of the platform object offer comprehensive and macroscopic perspectives. Its primary purpose is to establish abstract connections that integrate features of seller and customer objects. This holistic feature is instrumental in ensuring authoring content aligns with the platform's characteristics as follows:

\labelitemii \texttt{Product correlation $<P_0>$}: This feature is derived from the product taxonomy, which encompasses the distinctions and associations among diverse products. The integration of this feature enhances e-commerce authoring by providing a comprehensive understanding of product semantics through coherent and official ground-truth taxonomy labels based on expert knowledge.

\labelitemii \texttt{Platform background $<P_1>$}: This feature pertains to the background knowledge of specific e-commerce platforms. It aids the authoring process by aligning linguistic habits and policy knowledge corresponding to the platform representation. The textual information reflected by the official blogs and Q\&A pairs serve as the carrier of this feature.

Specifying an integrated instruction set containing the tasks that interleave essential features can improve the generalization and scenario understanding capabilities of LLMs~\cite{longpre2023flan}. In the e-commerce scenario, a productive approach involves formulating inference tasks that align with downstream applications' requirements to cover features from the participated objects. We formulate practical tasks for interleaving the features of objects, including Ads Generation, Query-enhanced Title Rewriting, Product Classification, Purchase Intent Speculation, and General E-commerce Q\&A. Table~\ref{tab_instr_fomulation} shows the instantiated instructions with highlighted object features.

\begin{table*}[htbp]
\centering
\caption{Examples of the instantiated instructions in e-commerce authoring interaction scenarios, where the tasks cover the object-interleaved features from \colorbox{pink}{seller \textit{<S>}}, \colorbox{lime}{customer \textit{<C>}}, and \colorbox{cyan}{platform \textit{<P>}}}
\label{tab_instr_fomulation}
\begin{adjustbox}{width=1\linewidth}
\begin{tabular}{l|l|l}
\hline
Task    & Instantiation    & Instruction \\ \hline
Ads Generation  &  \textit{<S>}  & Generate a short advertisement for the following product: \colorbox{pink}{[\texttt{product title}]}                    \\ \hline
Query-enhanced Title Rewriting &  \textit{<S, C$_0$>}  & Rewrite the product title of \colorbox{pink}{[\texttt{\texttt{product title}}]} according to the following query: \colorbox{lime}{[\texttt{query}]}.        \\ \hline
Product Classification &  \textit{<S, P$_0$>}  & What is the \colorbox{cyan}{[\texttt{product category}]} of this following product belongs to? \colorbox{pink}{[\texttt{product title}]}   \\ \hline
Purchase Intent Speculation &  \textit{<C$_1$, P$_0$>}  & Given the query of \colorbox{lime}{[\texttt{query}]}, which of the following \colorbox{cyan}{[\texttt{product category}]} is the customer interested in?                             \\ \hline
General E-commerce Q\&A & \texttt{<P$_1$>}   & \colorbox{cyan}{[\texttt{How am I charged for Ads?}]}                                                                              \\ \hline
\end{tabular}
\end{adjustbox}
\end{table*}

The ads generation aims to create compelling content highlighting product features and incorporating persuasive language to stimulate purchasing. This is the most prevalent task in e-commerce authoring. The query-enhanced title rewriting focuses on personalizing the original product titles for preference elicitation based on user queries, making them more appealing and aligned with user purchasing inclinations. Since the semantic features in product titles and queries have domain prominence, we utilize the product classification and purchase intention speculation tasks to map these semantic features to unified product taxonomy, shifting the model's focus from general to e-commerce knowledge. The purchase intention speculation also establishes semantic associations between queries and taxonomy to enhance query understanding and product recommendation. The general e-commerce Q\&A introduces background knowledge through Q\&A pairs defined by the platform. Its testing scenario can be seen as a zero-shot learning task for the platform style alignment. The formulated seed set consisting 300 instructions covering these tasks.

\subsection{Instruction Expansion}

To enhance the generalizability of LLaMA-E models in various downstream authoring applications, we utilize the GPT-3.5-turbo-0301 model as a teacher to expand the initial set of instructions. The expansion process involves rewriting the seed instructions with the teacher model to achieve a variety of expressions while maintaining semantic integrity. For tasks like product classification and intention speculation, where responses are strictly predefined, only the instructions are rewritten to maintain the necessary response constraints following the Prompt~\ref{prompt:prompt_inst_rew}. The \texttt{<seed instructions>} represents the raw expert-defined instructions.

\begin{promptbox}[prompt_inst_rew]{\normalsize Instruction Expansion}
  \small \textbf{[INST]} \texttt{Rewrite the following instruction while maintaining semantic consistency:}
  \textbf{[/INST]} \texttt{<seed instructions>}
\end{promptbox}

For generative tasks that encourage the production of varied linguistic expressions, we not only expand the raw instructions but also adopt two strategies to expand the responses corresponding to the instructions: response generation and rewriting. The response generation strategy utilizes the teacher model to generate appropriate responses based on the expanded instructions, thereby diversifying the raw responses by leveraging the parameterized knowledge encapsulated within LLMs. The prompt format is as shown in Prompt~\ref{prompt:prompt_resp_gen}, where \texttt{<expanded instructions>} represents the generative tasks' instructions that are expanded by the Prompt~\ref{prompt:prompt_inst_rew}, and \texttt{<seed inputs>} is the fixed authoring features like product title, taxonomy, etc.

\begin{promptbox}[prompt_resp_gen]{\normalsize Responses Generation}
  \small \textbf{[INST]} \texttt{<expanded instructions>}
  \textbf{[/INST]}\\ \texttt{<seed inputs>}
\end{promptbox}

The response rewriting strategy involves enabling the teacher model to rewrite the responses (as \texttt{<responses>}), thereby producing more diverse expressions while maintaining alignment with the fixed corresponding instructions paired. The prompt format is as depicted in Prompt~\ref{prompt:prompt_resp_rew}.

\begin{promptbox}[prompt_resp_rew]{\normalsize Responses Rewriting}
  \small \textbf{[INST]} \texttt{Rewrite the following generated response to diversify its expression:}
  \textbf{[/INST]} \texttt{<responses>}
\end{promptbox}

After generating the expanded set of instructions, a post-processing phase is conducted by domain experts. During this crucial phase, instruction-response pairs with duplicate content are filtered out to ensure the uniqueness and quality of the final instruction set. The refined instructions are then evenly distributed across the respective tasks, resulting in a comprehensive set of 120k instructions. This final instruction set is subsequently utilized to train the LLaMA-E models. The examples of the final instructions are provided in Appendix~\ref{sec_ei}.

\subsection{Instruction Tuning}

The LLaMA-E models are developed by integrating the proposed e-commerce authoring instruction set with LLaMA~\cite{touvron2023llama} models, utilizing parameter scales of 7b, 13b, and 30b as the base models. Deploying these large-parameter LLaMA models in customer-specific scenarios poses significant challenges due to the associated computational complexity. To address this, we employ LoRA~\cite{hu2021lora}, a Parameter-Efficient Fine-Tuning (PEFT) strategy that facilitates cost-effective fine-tuning while achieving results comparable to full model fine-tuning. LoRA is designed for low-rank adaptation, which reduces the number of trainable parameters in the fine-tuning process by learning rank-decomposition matrix pairs while keeping the original weights static. This method significantly enhances the LLaMA-E models' applicability in e-commerce authoring tasks, enabling the LLMs to effectively serve the scenario objects (particularly sellers and customers) even with limited computational resources. In our fine-tuning process, the forward pass of a linear layer represented by $h=W_0x$ in the base LLaMA models is modified with the LoRA. The process is described as Eq.\ref{eq_lora}.

\begin{equation}
\label{eq_lora}
h=W_0x+BAx,
\end{equation}
where $W_0\in{\mathbb{R}}^{d\times k}$ represents the frozen pre-trained weight matrices from the base LLaMA models, whereas $B\in{\mathbb{R}}^{d\times r}$ and $A\in{\mathbb{R}}^{r\times k}$ are the trainable parameters that are initialized with zero and Gaussian initialization, respectively. All variables with the rank of $min(d,k)$. Since LLaMA models are trained on general corpora like Wikipedia and C4, it is crucial to specifically align the focus towards comprehending unique e-commerce semantic features, such as rare stylistic words (e.g., Boho, Berber), when employing the models for authoring tasks. This emphasis is particularly important for modelling product descriptions and personalized customer queries. To enable the model to fit these nuanced features, we utilize LoRA with trainable parameters of $W_q$, $W_k$, $W_v$, and $W_o$, which are the weight matrices in the self-attention module.

\section{Experiment}

\subsection{Implementation Details}

The dataset for constructing the instruction set is sourced from practical e-commerce scenarios, featuring vital details of product titles, taxonomy, and customer queries. It also includes an action element reflecting customer interactions with the retrieved products, which has the value of "no action", "click", "cart add", and "purchase". To ensure the data reflects potential purchase interest based on the correlation between queries and products, data labelled as "no action" (indicating no interest) is filtered. The screened data undergoes post-processing to remove emojis and interfering characters.

The test set comprises 19,367 unseen product instances, each featuring an additional product description element for more detailed information. To evaluate the LLaMA-E models on general e-commerce Q\&A tasks, we utilized 30 authentic Q\&A pairs from the platform's "Help Center" that are not included in the training set. The LLaMA-E models are trained using two NVIDIA A40 GPUs. The number of trainable parameters and the training time per epoch are detailed in Table~\ref{tab_para}.

\begin{table}[htbp]
  \centering
  \caption{Training details of the LLaMA-E models}
  \label{tab_para}
  \begin{adjustbox}{width=1\linewidth}
  \begin{tabular}{c|c|c}
    \hline
    {Model} & {Trainable Parameters} & {GPU Hours} \\
    \hline
    LLaMA-E-7b & 8.39m & 3.93 \\
    \hline
    LLaMA-E-13b & 13.11m & 9.51 \\
    \hline
    LLaMA-E-30b & 25.56m & 41.14\\
    \hline
  \end{tabular}
  \end{adjustbox}
\end{table}

\subsection{Evaluation System}

The evaluation system is designed to assess the generalization capability of the LLaMA-E models in practical e-commerce applications. This assessment necessitates that the generated content prioritizes the coverage of essential features following the task requirements rather than toughly adhering to fixed responses based on instructions. The metrics in the evaluation system are as follows:

\textbf{Ads Generation}: The evaluation metrics for this task include $BLEU$~\footnote{\url{https://pypi.org/project/rouge/\#/}}~\cite{papineni2002bleu} and $ROUGE$~\footnote{\url{https://www.nltk.org/api/nltk.translate.bleu_score.html\#/}}~\cite{lin2004rouge}, which are commonly used in combination in NLG tasks~\cite{narasimhan2022towards}. We calculate the $BLEU$ and $ROUGE-L$ scores between the generated ads and the product title and description separately, denoted as $BL_{Ad_t}$, $BL_{Ad_d}$, $RL_{Ad_t}$, and $RL_{Ad_d}$. This evaluation aligns with the motivation of seller-written advertisements, assessing whether the generated contents incorporate the essential features in the titles and the significant details in the descriptions.

\textbf{Query-enhanced Title Rewriting}: We calculate the $BLEU$ and $ROUGE-L$ scores between the rewritten title and the raw product title and customer query separately, represented as $BL_{T_t}$, $RL_{T_t}$, $BL_{T_q}$, and $RL_{T_q}$. These metrics measure how comprehensively the rewritten title covers features from the raw title and query. The raw product titles are short sentences with dysfluent text stacked with discrete entities, making the readability a criterion for evaluating whether the rewritten title can be used in publicity scenes like banners. We calculate the perplexity ($PPL$)~\cite{jelinek1977perplexity} metric of the rewritten title by taking the GPT-2-XL~\footnote{\url{https://huggingface.co/gpt2-xl}} as the evaluation model, which boasts 1.5 billion parameters and is pre-trained on the WebText dataset with extensive general semantic features.

\textbf{Product Classification}: This task evaluates whether the LLaMA-E model can accurately classify products according to a predefined taxonomy based solely on their raw textual titles. The evaluation metrics include the macro-average Precision ($P_{pt}$), Recall ($R_{pt}$), and F1-score ($F_{1_{pt}}$).

\textbf{Intent Speculation}: This task evaluates the performance of the LLaMA-E in analyzing customer potential purchasing interest expressed by queries associated with the product taxonomy. The evaluation can be quantitatively measured using the classification metrics, including macro-average Precision ($P_{qs}$), Recall ($R_{qs}$), and F1-score ($F_{1_{qs}}$).

\textbf{General E-commerce Q\&A}: The metrics of $BLEU$ and $ROUGE-L$ measure the explicit overlap and similarity between the generated and standard answers for evaluating the generalization on unseen questions, represented as $BL_{qa}$ and $RL_{qa}$. We also introduce the average BERT Score ($BE_{qa}$)~\cite{bert-score} for evaluating the implicit semantic similarity. This can be regarded as a measure of the platform-specific knowledge conveyed by semantics injected into the LLMs.

\textbf{Overall}: We calculate an overall metric, the geometric mean ($GM$)~\cite{yi2020texts}, of all the aforementioned evaluation metrics to assess model performance comprehensively. The $PPL$ metric is transformed to $\frac{1}{\ln{PPL}}$ for the calculation to comply with the monotonicity of the $GM$ metric, where a higher $GM$ indicates better overall performance.

\subsection{Baseline Methods}

We compare the proposed LLaMA-E models with the LLMs of \textbf{GPT-2}~\cite{radford2019language}, \textbf{BART}~\cite{lewis2020bart}, \textbf{T5-base}~\cite{Raffel2020Exploring}, \textbf{GPT-Neo}~\cite{gpt-neo}, and \textbf{LLaMA}~\cite{touvron2023llama}. We use the proposed instruction set to fine-tune GPT-2 and BART models for each authoring task. This evaluation examines the distinction between comprehensive instruction fine-tuning and separate task-specific fine-tuning when conducting correlated tasks under the same scenario. The other baselines are introduced with their pre-trained general models, and the comparison of the LLaMA-7/13/30b models can be regarded as the ablation study to evaluate the advantages of designed object-interleaved instructions in enabling general LLMs to learn e-commerce authoring knowledge. Additionally, we report the performance of the teacher model, GPT-3.5-turbo-0310, on each of the evaluation tasks.

\section{Result and Analysis}

\subsection{Quantitative Evaluation}

The quantitative evaluation results are shown in Table~\ref{tab_quan}, and we also show the qualitative evaluation results in Appendix~\ref{sec:qual_eval}. The LLaMA-E models have generally achieved better results than the baselines in most quantitative metrics. The LLaMA-E-7b model significantly outperforms other baselines in the $GM$ metric, proving it has the best overall performance in the required e-commerce authoring tasks. Within the internal comparison of the LLaMA-E models, a significant trend is the gradual enhancement of performance in classification tasks as the scale of parameters increases. This demonstrates that a larger parameter scale helps fit the more granular scenario features within the instruction set. However, one potential drawback is overfitting, stemming from the limitations in the scale and diversity of the current instruction set. The models' ability to generalize knowledge from general corpora and effectively model natural language may be affected. This observation also confirms that smaller-scale models may be adequate for tasks with lower inference requirements, thereby avoiding complex computational costs.

\begin{table*}[htb]
\centering
\caption{The quantitative evaluation results of the LLaMA-E models and baselines, where the best results are \textbf{bolded} and the second best are \underline{underlined}. The model achieves the highest $GM\uparrow$ metric is \colorbox{green!20}{highlighted}}
\label{tab_quan}
\begin{adjustbox}{width=1\linewidth}
\begin{tabular}{cc|cccc|ccccc|ccc|ccc|ccc|c}
\hline
\multicolumn{2}{c|}{\multirow{2}{*}{Model}}       & \multicolumn{4}{c|}{Ads Generation}	& \multicolumn{5}{c|}{Query-enhanced Title Rewriting}	&	\multicolumn{3}{c|}{Product Classification}	& \multicolumn{3}{c|}{Intent Speculation}	& \multicolumn{3}{c|}{General Q\&A}	& \multirow{2}{*}{$GM\uparrow$} \\ \cline{3-20}
\multicolumn{2}{c|}{}                             & \multicolumn{1}{c|}{$BL_{A_t}$}		& \multicolumn{1}{c|}{$RL_{A_t}$} 		& \multicolumn{1}{c|}{$BL_{A_d}$} 			& \multicolumn{1}{c|}{$RL_{A_d}$} 		& \multicolumn{1}{c|}{$BL_{T_t}$} & \multicolumn{1}{c|}{$RL_{T_t}$} & \multicolumn{1}{c|}{$BL_{T_q}$} & \multicolumn{1}{c|}{$RL_{T_q}$} & $PPL$	& \multicolumn{1}{c|}{$P_{pt}$} & \multicolumn{1}{c|}{$R_{pt}$} & $F_{1_{pt}}$	& \multicolumn{1}{c|}{$P_{qs}$} & \multicolumn{1}{c|}{$R_{qs}$} & $F_{1_{qs}}$	& \multicolumn{1}{c|}{$BL_{qa}$} 	& \multicolumn{1}{c|}{$RL_{qa}$} & $BE_{qa}$ &                              \\ \hline \hline
\multicolumn{2}{c|}{GPT-3.5}                      & \multicolumn{1}{c|}{16.76}            & \multicolumn{1}{c|}{47.65} & \multicolumn{1}{c|}{0.56}    & \multicolumn{1}{c|}{11.15}           & \multicolumn{1}{c|}{26.08}            &  \multicolumn{1}{c|}{60.04}           & \multicolumn{1}{c|}{9.10}          & \multicolumn{1}{c|}{35.00}          & 120.86	& \multicolumn{1}{c|}{49.48}         & \multicolumn{1}{c|}{49.23}         & 49.35	& \multicolumn{1}{c|}{19.58}         & \multicolumn{1}{c|}{19.18}         & 19.38             & \multicolumn{1}{c|}{2.83}          & \multicolumn{1}{c|}{14.41}     &   85.53  & 15.06                             \\ \hline \hline
\multicolumn{2}{c|}{GPT-2}                        & \multicolumn{1}{c|}{\underline{14.85}}       & \multicolumn{1}{c|}{25.03} & \multicolumn{1}{c|}{0.29} & \multicolumn{1}{c|}{6.83}            & \multicolumn{1}{c|}{16.57}            & \multicolumn{1}{c|}{39.48}            & \multicolumn{1}{c|}{1.64}          & \multicolumn{1}{c|}{19.98}          & 253.73	& \multicolumn{1}{c|}{\textbf{87.50}}         & \multicolumn{1}{c|}{24.01}         & 33.18  & \multicolumn{1}{c|}{56.25}         & \multicolumn{1}{c|}{6.33}         & 10.69             & \multicolumn{1}{c|}{2.14}          & \multicolumn{1}{c|}{11.42}     &  85.66   & 10.26                             \\ \hline
\multicolumn{2}{c|}{BART}                         & \multicolumn{1}{c|}{13.05}  & \multicolumn{1}{c|}{36.04}  & \multicolumn{1}{c|}{0.37}            & \multicolumn{1}{c|}{8.37}            & \multicolumn{1}{c|}{18.64}            &  \multicolumn{1}{c|}{41.40}           & \multicolumn{1}{c|}{5.75}          & \multicolumn{1}{c|}{20.33}          & \multicolumn{1}{c|}{389.35}        & \multicolumn{1}{c|}{73.75}         & \multicolumn{1}{c|}{54.82}         & 62.39 & \multicolumn{1}{c|}{66.67}         & \multicolumn{1}{c|}{47.97}         & 54.71             & \multicolumn{1}{c|}{\underline{3.32}}          & \multicolumn{1}{c|}{\underline{14.02}}     &  86.02   & 15.83                             \\ \hline \hline
\multicolumn{2}{c|}{T5-base}                      & \multicolumn{1}{c|}{14.55}            & \multicolumn{1}{c|}{37.96} & \multicolumn{1}{c|}{\underline{0.92}} & \multicolumn{1}{c|}{9.10}           & \multicolumn{1}{c|}{\underline{21.16}}            & \multicolumn{1}{c|}{\underline{53.42}}            & \multicolumn{1}{c|}{\textbf{7.95}}          & \multicolumn{1}{c|}{23.82}          & 300.02 & \multicolumn{1}{c|}{40.04}         & \multicolumn{1}{c|}{9.52}         & 9.62 & \multicolumn{1}{c|}{26.17}         & \multicolumn{1}{c|}{9.98}         & 9.01              & \multicolumn{1}{c|}{3.25}          & \multicolumn{1}{c|}{13.99}     &   85.33  & 11.03                             \\ \hline
\multicolumn{2}{c|}{GPT-Neo}                       & \multicolumn{1}{c|}{12.93}            & \multicolumn{1}{c|}{30.62}    & \multicolumn{1}{c|}{\textbf{0.97}} & \multicolumn{1}{c|}{8.16}            & \multicolumn{1}{c|}{\textbf{21.43}}            & \multicolumn{1}{c|}{49.04}            & \multicolumn{1}{c|}{\underline{7.21}}          & \multicolumn{1}{c|}{\underline{25.49}}          & 306.83 & \multicolumn{1}{c|}{9.88}         & \multicolumn{1}{c|}{5.86}         & 2.42 & \multicolumn{1}{c|}{2.61}         & \multicolumn{1}{c|}{5.05}         & 1.61             & \multicolumn{1}{c|}{2.41}          & \multicolumn{1}{c|}{10.10}     &  83.56   & 6.65                             \\ \hline
\multicolumn{1}{c|}{\multirow{3}{*}{\rotatebox{90}{\small{LLaMA}}}} & 7b  &  \multicolumn{1}{c|}{10.05}    & \multicolumn{1}{c|}{21.63}     & \multicolumn{1}{c|}{0.77}            & \multicolumn{1}{c|}{8.52}            & \multicolumn{1}{c|}{12.00}            & \multicolumn{1}{c|}{27.32}            & \multicolumn{1}{c|}{3.22}          & \multicolumn{1}{c|}{13.86}          & 206.71 & \multicolumn{1}{c|}{28.64}         & \multicolumn{1}{c|}{4.29}         & 4.12  & \multicolumn{1}{c|}{9.64}         & \multicolumn{1}{c|}{3.01}         & 2.29             & \multicolumn{1}{c|}{2.01}          & \multicolumn{1}{c|}{11.17}    &   84.81   &  6.31                            \\ \cline{2-21} 
\multicolumn{1}{c|}{}                       & 13b & \multicolumn{1}{c|}{6.31}            & \multicolumn{1}{c|}{16.35}  & \multicolumn{1}{c|}{0.75}   & \multicolumn{1}{c|}{7.94}           & \multicolumn{1}{c|}{15.28}            &  \multicolumn{1}{c|}{30.40}           & \multicolumn{1}{c|}{3.35}          & \multicolumn{1}{c|}{13.61}          & 181.54 & \multicolumn{1}{c|}{19.64}         & \multicolumn{1}{c|}{1.78}         & 2.62 & \multicolumn{1}{c|}{13.62}         & \multicolumn{1}{c|}{3.48}         & 4.79             & \multicolumn{1}{c|}{0.86}          & \multicolumn{1}{c|}{11.53}    &   84.39   & 5.72                             \\ \cline{2-21} 
\multicolumn{1}{c|}{}                       & 30b & \multicolumn{1}{c|}{12.67}            & \multicolumn{1}{c|}{22.93}  & \multicolumn{1}{c|}{0.91}   & \multicolumn{1}{c|}{7.44}           & \multicolumn{1}{c|}{18.03}            & \multicolumn{1}{c|}{32.03}            & \multicolumn{1}{c|}{3.15}          & \multicolumn{1}{c|}{12.95}          & 159.18 & \multicolumn{1}{c|}{32.15}         & \multicolumn{1}{c|}{6.12}         & 9.27 & \multicolumn{1}{c|}{11.54}         & \multicolumn{1}{c|}{4.25}         & 5.73             & \multicolumn{1}{c|}{2.49}          & \multicolumn{1}{c|}{11.38}     &  84.55   & 7.79                             \\ \hline \hline
\multicolumn{1}{c|}{\multirow{3}{*}{\rotatebox{90}{{\small LLaMA-E}}}} & \cellcolor{green!20} 7b  & \multicolumn{1}{c|}{\cellcolor{green!20}\textbf{15.18}}            & \multicolumn{1}{c|}{\cellcolor{green!20}46.96} & \multicolumn{1}{c|}{\cellcolor{green!20}0.45} & \multicolumn{1}{c|}{\cellcolor{green!20}\underline{9.87}}            & \multicolumn{1}{c|}{\cellcolor{green!20}18.88}            & \multicolumn{1}{c|}{\cellcolor{green!20}\textbf{54.36}}            & \multicolumn{1}{c|}{\cellcolor{green!20}4.66}          &  \multicolumn{1}{c|}{\cellcolor{green!20}\textbf{25.69}}          & \cellcolor{green!20} \textbf{132.86} & \multicolumn{1}{c|}{\cellcolor{green!20}60.03}         & \multicolumn{1}{c|}{\cellcolor{green!20}63.80}         & \cellcolor{green!20} 59.01  & \multicolumn{1}{c|}{\cellcolor{green!20}59.52}         & \multicolumn{1}{c|}{\cellcolor{green!20}61.09}         &      \cellcolor{green!20}  59.71              & \multicolumn{1}{c|}{\cellcolor{green!20}\textbf{4.04}}          & \multicolumn{1}{c|}{\cellcolor{green!20}\textbf{15.86}}     &  \cellcolor{green!20} \textbf{86.43}  & \cellcolor{green!20} \textbf{17.41}                             \\ \cline{2-21} 
\multicolumn{1}{c|}{}                       & 13b & \multicolumn{1}{c|}{13.08}            & \multicolumn{1}{c|}{\underline{46.99}}  & \multicolumn{1}{c|}{0.32} & \multicolumn{1}{c|}{8.99}           & \multicolumn{1}{c|}{15.07}            &  \multicolumn{1}{c|}{50.48}           & \multicolumn{1}{c|}{4.15}          & \multicolumn{1}{c|}{23.21}          & \underline{152.23} & \multicolumn{1}{c|}{72.51}         & \multicolumn{1}{c|}{\underline{68.92}}         & \underline{69.99} & \multicolumn{1}{c|}{\underline{72.87}}         & \multicolumn{1}{c|}{\underline{68.08}}         &  \underline{69.62}             & \multicolumn{1}{c|}{\underline{3.32}}          & \multicolumn{1}{c|}{12.36}     &  \underline{86.14}   & 16.77                           \\ \cline{2-21} 
\multicolumn{1}{c|}{}                       & 30b & \multicolumn{1}{c|}{14.23}            & \multicolumn{1}{c|}{\textbf{47.23}}  & \multicolumn{1}{c|}{0.41} & \multicolumn{1}{c|}{\textbf{10.32}}           & \multicolumn{1}{c|}{15.96}            & \multicolumn{1}{c|}{52.95}            & \multicolumn{1}{c|}{4.27}          & \multicolumn{1}{c|}{24.60}          & 177.75        & \multicolumn{1}{c|}{\underline{74.32}}         & \multicolumn{1}{c|}{\textbf{73.16}}         &  \textbf{71.75} & \multicolumn{1}{c|}{\textbf{74.51}}         & \multicolumn{1}{c|}{\textbf{72.18}}         & \textbf{70.53}            & \multicolumn{1}{c|}{2.28}          &  \multicolumn{1}{c|}{13.29}    & 86.01    & \underline{17.28}                             \\ \hline 
\end{tabular}
\end{adjustbox}
\end{table*}

Compared to the teacher model, GPT-3.5, the LLaMA-E model achieved competitive performance in the $BLEU$ and $ROUGE-L$ metrics, which evaluate text overlap, as well as in the $PPL$ metric, which assesses model performance in generating qualified text. These results indicate that the text generated by the LLaMA-E models aligns closely with GPT-3.5 in terms of information coverage and readability. In tasks such as product classification, intent speculation, and general e-commerce Q\&A—each requiring professional domain knowledge—the LLaMA-E models demonstrated superior performance. This underscores that general LLMs are not yet sufficient to meet the fine-grained requirements of domain-specific applications, highlighting the necessity of designing LLMs tailored to scenario features. This comparison validates the feasibility of aligning general LLMs to focus on practical downstream e-commerce authoring applications through specially designed instructions that comprehend object-interleaved features.

Compared to the task-specific fine-tuned GPT-2 and BART models, the LLaMA-E models achieved superior performance in the $GM$ metric, demonstrating that the designed instruction set provides a more fine-grained fit to the comprehensive features of the given tasks than task-specific fine-tuning. Both of these two models outperformed the remaining baselines in the $F_{1_{qs}}$, $F_{1_{pt}}$, and $BE_{qa}$ metrics, with GPT-2 even achieving the best performance in the $P_{pt}$ metric. These results indicate that incorporating domain knowledge significantly enhances the serviceability of LLMs in specific scenarios. However, these models require cumbersome task-isolated fine-tuning, and their limited in-context learning ability further restricts the efficient utilization of the features in available training data. These limitations hinder their practical applicability.

The remaining baselines, T5-base, GPT-Neo, and LLaMA, which are of similar scale to the LLaMA-E models, are incorporated to evaluate the applicability of extensive general knowledge in specific application scenarios. The findings indicate that while these models excel in certain generative metrics, they fall short in classification and $BE_{qa}$ metrics. This suggests that large-scale general knowledge enables these models to parameterize the basic linguistics features to generate readable yet context-independent text, limiting their ability to represent fine-grained scenario-specific knowledge and provide precise support for e-commerce authoring services. This hypothesis will be further examined in qualitative evaluations in Appendix~\ref{sec:qual_eval}. Compared to the LLaMA models for ablation studies, the LLaMA-E models perform better across all metrics, validating the positive support of the proposed instruction set for authoring scenarios.

\subsection{Human Evaluation}

\begin{figure}[htbp]
 \centering
 \includegraphics[width=0.46\textwidth]{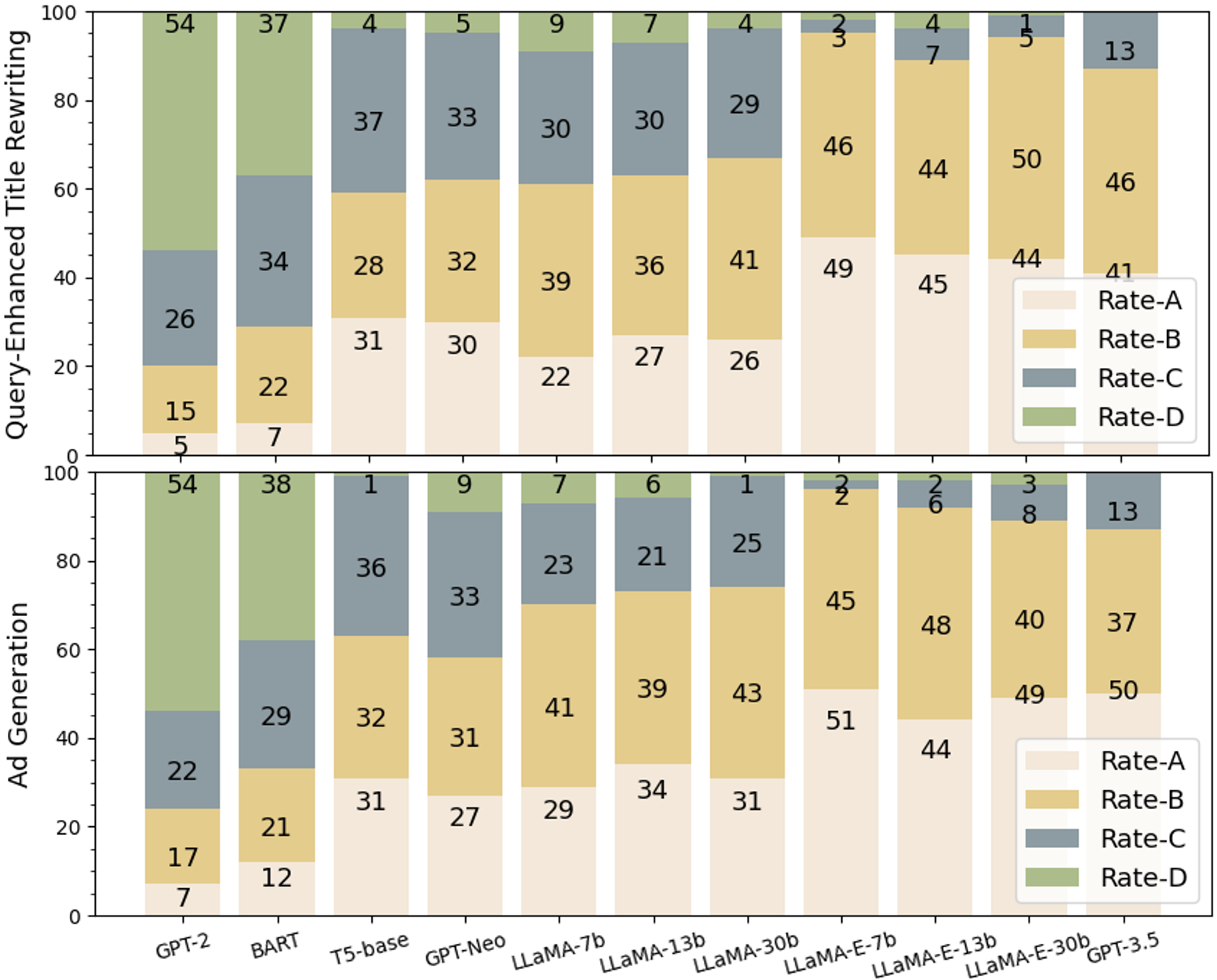}
 \caption{Human evaluation of ads generation and query-enhanced title rewriting. The legend marks rating}
 \label{fig:human_eval}
\end{figure}

We invited ten volunteer annotators with extensive experience in English comprehension and e-commerce to conduct human evaluations on tasks of ads generation and query-enhanced title rewriting. Each annotator is asked to anonymously rate ten randomly selected generated texts from the progressive perspectives of readability, coverage, and attractiveness based on the following criteria:

\begin{itemize}
\item \texttt{Rate-A}: The generated text captivates customers and encourages purchases while covering essential features of products or queries.
\item \texttt{Rate-B}: The generated text covers the essential features of products or queries but lacks attractiveness or persuasive appeal in stylistic.
\item \texttt{Rate-C}: The generated text is legible and presented in fluent natural language, but some essential features in the input are lost.
\item \texttt{Rate-D}: The generated text cannot be understood due to issues such as messy syntax.
\end{itemize}

Figure~\ref{fig:human_eval} illustrates the results of human evaluation, showing the rating distribution of the test samples. The findings indicate that the LLaMA-E models achieve competitive rating scores compared to the GPT-3.5 while outperforming other baseline models. Additionally, the annotators report that the title rewriting outputs generated by the LLaMA-E models are more attractive. This observed advantage can be primarily attributed to the ads generation task during the instruction fine-tuning process, which encourages the model to produce captivating phrases like "order now" and other persuasive language. This phenomenon indicates the beneficial correlation within the instruction set. The GPT-3.5 shows stronger robustness as it generates text without unreadable content, proving the stability potential of larger LLMs in practical applications.

\section{Conclusion and Future Research}

This paper proposes and releases LLaMA-E, the LLMs toiled for e-commerce authoring. To comprehensively understand e-commerce scenarios, the instructions prioritize integrating interleaved features presented by essential participated objects derived from practical tasks, aligning the LLMs' focus with e-commerce-specific knowledge. This approach provides a paradigm that emphasizes the crucial role of fine-grained scenario object features for aligning the general LLMs to the domain-specific applications, suggesting an inspirational solution for empowering diverse downstream LLMs-based inference tasks. Compared with other baselines, the LLaMA-E models achieve state-of-the-art results in comprehensive evaluation systems.

Diversifying existing models to cover a wider range of authoring tasks is crucial for future research. Moreover, extending the models to operate in multilingual environments is essential for providing more extensive services. Certain research endeavours focus on dynamically injecting real-time knowledge into LLMs through information retrieval. Investigating personalized retrieval methods to incorporate preference features into e-commerce authoring models is also appealing, which can reduce tuning costs and promote more customized and private content-based applications.

\section*{Limitations}

While the proposed LLaMA-E models demonstrate promising results in empowering e-commerce authoring, certain limitations should be acknowledged. The current LLaMA-E models are primarily based on English data, and their performance in other languages and cross-lingual settings remains unexplored. Adapting the models to multilingual scenarios is crucial for providing comprehensive and uniform e-commerce services globally, as most e-commerce platforms operate worldwide. Moreover, this paper focuses on generating textual content for authoring purposes. However, e-commerce platforms often involve multimodal data, such as images and videos. Extending the models to leverage multimodal information could yield more engaging and informative authoring content. Although the paper emphasizes aligning LLMs with domain-specific knowledge, the models' capabilities in handling real-time updates or rapidly evolving trends in the e-commerce landscape are not thoroughly investigated. Mechanisms for continual learning and knowledge refreshments are beneficial for maintaining the models' relevance. The experiments are conducted on the NVIDIA A40 GPU servers. However, in practical e-commerce applications, some service users, such as sellers and customers, do not have server-level computing hardware. Therefore, further exploring the model's response performance on different PCs and optimizing the model scale accordingly is helpful for the widespread application of the model.

\bibliography{anthology, custom}

\newpage
\clearpage

\appendix

\setcounter{figure}{0}
\setcounter{table}{0}
\renewcommand{\thefigure}{A\arabic{figure}}
\renewcommand{\thetable}{A\arabic{table}}

\section{Expanded Instructions}
\label{sec_ei}
This appendix presents examples of the expanded instructions for the aforementioned e-commerce authoring tasks in Table~\ref{tab_ei}.

\begin{table}[htbp]
\centering
\caption{Examples of the expanded instructions for each of the e-commerce authoring tasks}
\label{tab_ei}
\begin{adjustbox}{width=1\linewidth}
\begin{tabular}{c|p{8cm}}
\hline
Task                                            & Expanded Instructions                                                                        \\ \hline
\multirow{7}{*}{\rotatebox{90}{\thead{Ads\\ Generation}}}                 & Produce an advertisement for the specified product.                                          \\ \cline{2-2} 
                                                & Create an advertisement for the specified product.                                           \\ \cline{2-2} 
                                                & Produce an advertisement for the product mentioned below.                                    \\ \cline{2-2} 
                                                & Generate an ad designated for the following product.                              \\ \cline{2-2} 
                                                & Prepare an advertisement for the product provided below.                                     \\ \hline \hline
\multirow{10}{*}{\rotatebox{90}{\thead{Query-enhanced \\Title Rewriting}}} & Rephrase the subsequent product title along with the query.                                  \\ \cline{2-2} 
                                                & Revise the subsequent product title alongside the query.                                     \\ \cline{2-2} 
                                                & Revise the product title below, incorporating the given query.                               \\ \cline{2-2} 
                                                & Revise the given product title in combination with the query.                                \\ \cline{2-2} 
                                                & Incorporate the following query to rewrite the product title.                              \\ \hline \hline
\multirow{7}{*}{\rotatebox{90}{\thead{Product\\ Classification}}}         & To which category does the subsequent product belongs?                                         \\ \cline{2-2} 
                                                & Which category of this product belongs?                                          \\ \cline{2-2} 
                                                & Identify the category to which the following product belongs.                                \\ \cline{2-2} 
                                                & What category does the listed product belong to?                                              \\ \cline{2-2} 
                                                & Identify the category of the listed product.                                                \\ \hline \hline
\multirow{10}{*}{\rotatebox{90}{\thead{Purchase Intent\\ Speculation}}}    & Which category does the provided query imply the customer is interested in?                   \\ \cline{2-2} 
                                                & Based on the following query, which category does it indicate the customer is interested in? \\ \cline{2-2} 
                                                & What category is suggested by the following customer query's apparent interest?              \\ \cline{2-2} 
                                                & What category does the given query indicate the customer's interest in?                      \\ \cline{2-2} 
                                                & Identify the category that the following query suggests the customer is interested in.       \\ \hline \hline
\multirow{8}{*}{\rotatebox{90}{\thead{General\\ E-commerce Q\&A}}}                   & How are my orders attributed to Offsite Ads?                                                 \\ \cline{2-2} 
                                                & Describe the process of attributing my orders to Offsite Ads.                         \\ \cline{2-2} 
                                                & Can you explain how my orders are attributed to Offsite Ads?                                 \\ \cline{2-2} 
                                                & Please elaborate on the process of attributing my orders to Offsite Ads.                     \\ \cline{2-2} 
                                                & How are my orders linked to Offsite Ads?                                                     \\ \hline
\end{tabular}
\end{adjustbox}
\end{table}

\section{Qualitative Evaluation}
\label{sec:qual_eval}

This appendix presents the qualitative evaluation results for the tasks in the evaluation system.

\subsection{Ads Generation}

\begin{table}[htbp]
\centering
\caption{Qualitative results of ads generation}
\label{tab_qual_Ad}
\begin{adjustbox}{width=1\linewidth}
\begin{tabular}{cc|p{7.5cm}}
\hline
\multicolumn{2}{c|}{Model}                                                           & Generated Ads        \\ \hline \hline
\multicolumn{2}{c|}{\color{olive}Instr.}                                                          & \multicolumn{1}{l}{Generate an ad for the following product.} \\ \hline
\multicolumn{2}{c|}{\color{teal}Input}                                                          & \multicolumn{1}{p{6.6cm}}{Himalayan pink salt lamp bowl massage balls. Pure authentic. Dimmer switch. Night light. Home decor.} \\ \hline \hline
\multicolumn{2}{c|}{GPT-3.5}                                                         & Relax with pure Himalayan bliss. Authentic salt lamp bowl with massage balls. Dimmable night light.                     \\ \hline \hline
\multicolumn{2}{c|}{GPT-2}                                                           & Himalayan pink salt lamp bowl massage balls. Pure authentic. Pure authentic. Pure authentic...                    \\ \hline
\multicolumn{2}{c|}{BART}                                                            & Enhance ambiance with salt lamp bowl massage balls. Pure authentic ambiance. Perfect for home decor.                     \\ \hline \hline
\multicolumn{2}{c|}{T5-base}                                                         & Introducing our luxurious and stylish linen spherical pillow that offers an authentic, dimmer switch for a cozy night light in your home.                     \\ \hline
\multicolumn{2}{c|}{GPT-Neo}                                                          & Introducing our new Himalayan Pink Salt Lamp Bowl Massage Balls! You can feel the power of the natural light in your bedroom...                     \\ \hline
\multicolumn{1}{c|}{\multirow{5}{*}{\rotatebox{90}{\footnotesize{LLaMA}}}} & 7b  & Himalayan Pink Salt Lamp Bowl Massage Balls are perfect for any room in your house.                     \\ \cline{2-3} 
\multicolumn{1}{c|}{}                                                          & 13b & This light gives your room a warm glow and is made of real salt from Pakistan.                     \\ \cline{2-3} 
\multicolumn{1}{c|}{}                                                          & 30b & Himalayan Pink Lamp Bowl! I hope you like it :)                    \\ \hline \hline
\multicolumn{1}{c|}{\multirow{7}{*}{\rotatebox{90}{\footnotesize{LLaMA-E}}}} & \cellcolor{hiscolor}{7b}  & \cellcolor{hiscolor}{Purchase the Himalayan Pink Salt Lamp for a calming ambience and aesthetic. Buy now!}                    \\ \cline{2-3} 
\multicolumn{1}{c|}{}                                                          & 13b & Himalayan Salt Lamp Bowl: Relaxing, Dimmable, Massage Balls. Pure Authentic. Order it for Home Decor.                     \\ \cline{2-3} 
\multicolumn{1}{c|}{}                                                          & 30b & Himalayan salt lamp bowl with massage balls and dimmer: night light, decor. Buy now!                     \\ \hline 
\end{tabular}
\end{adjustbox}
\end{table}

This qualitative evaluation for this task involves assessing if the model creates compelling advertisements highlighting the product features. The results are presented in Table~\ref{tab_qual_Ad}. A notable advantage of the LLaMA-E model is its ability to generate persuasive phrases like "Buy now", effectively stimulating purchases and enhancing the advertising impact of the generated text. A more profound content requirement for ads is promoting abstract features that resonate with the customer's purchasing psychology, such as LLaMA-E portraying the product as inducing a calming and relaxing effect. This phenomenon is absent in ads generated by other baselines, indicating their limited ability to effectively correlate other relevant features within the scenario when completing specific tasks. Moreover, text generated by T5-base and LLaMA-13b tends to hallucinate, whereas the GPT-2 model produces incomprehensible and repetitive words.

\subsection{Query-enhanced Title Rewriting}

\begin{table}[htbp]
\centering
\caption{Qualitative results of query-enhanced title rewriting}
\label{tab_qual_tw}
\begin{adjustbox}{width=1\linewidth}
\begin{tabular}{cc|p{7cm}}
\hline
\multicolumn{2}{c|}{Model}                                                           & Rewritten Title        \\ \hline \hline
\multicolumn{2}{c|}{\color{olive}Instr.}                                                          & \multicolumn{1}{p{6.7cm}}{Rewrite the following product title combined with the query of "birthday gift men"} \\ \hline
\multicolumn{2}{c|}{\color{teal}Input}                                                          & \multicolumn{1}{p{6.7cm}}{Vintage birthday shirt fiftieth gift ideas} \\ \hline \hline
\multicolumn{2}{c|}{GPT-3.5}                                                         & Vintage 50th birthday shirt: Unique gift ideas for men's milestone celebrations.                     \\ \hline \hline
\multicolumn{2}{c|}{GPT-2}                                                           & vintage birthday shirt fiftieth gift ideasth birthday gift men's shirt fiftieth gift ideasth birthday shirt fiftieth gift...                    \\ \hline
\multicolumn{2}{c|}{BART}                                                            & vintage birthday shirt fiftieth birthday gift ideas for gifting birthday gifts and more!                     \\ \hline \hline
\multicolumn{2}{c|}{T5-base}                                                         & Personalised Birthday Gift for Father's Birthday Gift Men                     \\ \hline
\multicolumn{2}{c|}{GPT-Neo}                                                          & Men's Vintage Birthday Gift Ideas                     \\ \hline
\multicolumn{1}{c|}{\multirow{6}{*}{\rotatebox{90}{\footnotesize{LLaMA}}}} & 7b  & The birthday man can get a gift from this vintage shirt as it's for his fiftieth birthday...                    \\ \cline{2-3} 
\multicolumn{1}{c|}{}                                                          & 13b & A good example of a birthday gift for a fifty-year-old man would be a customized T-shirt...                     \\ \cline{2-3} 
\multicolumn{1}{c|}{}                                                          & 30b & Fifty Birthday Shirts For Men, Unique Funny T-Shirts With Slogans For The Big Five-Oh!                     \\ \hline \hline
\multicolumn{1}{c|}{\multirow{7}{*}{\rotatebox{90}{\footnotesize{LLaMA-E}}}} & \cellcolor{hiscolor}{7b}  & \cellcolor{hiscolor}{Vintage Shirt for Men's Birthday Gift, Ideal for Husbands, Dads, and Brothers.}     
\\ \cline{2-3} 
\multicolumn{1}{c|}{}                                                          & 13b & Vintage 50th Birthday Shirt, Perfect for Men. Customize with Name, Age, and Number. Ideal Gift.                     \\ \cline{2-3} 
\multicolumn{1}{c|}{}                                                          & 30b & Vintage Birthday Shirt - A Perfect Gift Idea for Men's Fiftieth Birthday. Shop Now.                   \\ \hline 
\end{tabular}
\end{adjustbox}
\end{table}

Query-enhanced title rewriting requires the rewritten title to contain the key features in both the original title and query. The results are in Table~\ref{tab_qual_tw}. The complex requirements of this task resulted in baselines rewriting titles that are not readable (GPT-2, BART, T5-base). The GPT-Neo rewritten title lost the crucial product feature of "shirt". The LLaMA models cover titles and queries by generating redundant descriptors, indicating misunderstandings of instructions and factual deviation. In contrast, the titles rewritten by LLaMA-E models cover key features, while LLaMA-7b deduces and visualizes the "fiftieth" in "Input" as a male character of the corresponding age group, showcasing chain-of-thought reasoning ability leveraging domain-specific knowledge.

\subsection{Purchase Intent Speculation}

\begin{table}[htbp]
\centering
\caption{Qualitative results of purchase intent speculation}
\label{tab_qual_is}
\begin{adjustbox}{width=1\linewidth}
\begin{tabular}{cc|p{7cm}}
\hline
\multicolumn{2}{c|}{Model}                              & Purchase Intent  \\ \hline \hline
\multicolumn{2}{c|}{\color{olive}Instr.}                & \multicolumn{1}{p{6.6cm}}{Which category does the following query suggest the user is interested in?} \\ \hline
\multicolumn{2}{c|}{\color{teal}Input}                  & \multicolumn{1}{p{6.6cm}}{Himalayan salt lamp} \\ \hline 
\multicolumn{2}{c|}{\color{orange}Taxonomy}                  & \multicolumn{1}{p{6.6cm}}{home and living} \\ \hline \hline
\multicolumn{2}{c|}{GPT-3.5}                                                         & Home and Living                     \\ \hline \hline
\multicolumn{2}{c|}{GPT-2}                              & jewelry                         \\ \hline
\multicolumn{2}{c|}{\cellcolor{hiscolor}{BART}}                               &  \cellcolor{hiscolor}{home and living}                        \\ \hline \hline
\multicolumn{2}{c|}{T5-base}                            & The query suggests that the user is interested in "toys and games" category.
                         \\ \hline
\multicolumn{2}{c|}{GPT-Neo}                            & The query suggests that the user is interested in accessories, specifically a "Himalayan salt lamp".                         \\ \hline
\multicolumn{1}{c|}{\multirow{5}{*}{\rotatebox{90}{\footnotesize{LLaMA}}}} & 7b  & This query shows the products with Himalayan salt lamp as part of their title or description on Amazon.com.                         \\ \cline{2-3} 
\multicolumn{1}{c|}{}                                                      & 13b & https://www.etsy.com/search?q=himalayan+salt+lamp                         \\ \cline{2-3} 
\multicolumn{1}{c|}{}                                                      & 30b & Home \& Living                         \\ \hline \hline
\multicolumn{1}{c|}{\multirow{3}{*}{\rotatebox{90}{\footnotesize{LLaMA-E}}}} & \cellcolor{hiscolor}{7b}  & \cellcolor{hiscolor}{home and living}                         \\ \cline{2-3} 
\multicolumn{1}{c|}{}                                                        & \cellcolor{hiscolor}{13b} & \cellcolor{hiscolor}{home and living}                         \\ \cline{2-3} 
\multicolumn{1}{c|}{}                                                        & \cellcolor{hiscolor}{30b} & \cellcolor{hiscolor}{home and living}                         \\ \hline
\end{tabular}
\end{adjustbox}
\end{table}

This evaluation can be achieved by measuring whether the generated intention contains the corresponding taxonomy keywords based on the given query. The results are shown in Table~\ref{tab_qual_is}. The fine-tuned LLMs (BART, GPT-2, and LLaMA-E) can accurately identify specific categories from the product taxonomy. Although GPT-2 incorrectly inferred the customer's purchase intention as "jewelry", this category still falls within the standard taxonomy. In contrast, due to the lack of domain-specific knowledge and adherence to instruction constraints, the intention generated by general LLMs is diverse and unrelated to the instructions, which limits their serviceability in practical applications. This phenomenon supports the hypothesis in quantitative evaluations that such models have better text generation metrics but poorer classification metrics. Despite GPT-3.5 performing better than other baselines, it still exhibits the aforementioned issues in other testing cases. This emphasizes the necessity of domain knowledge to enhance the applicability of LLMs to specific scenarios.

\subsection{General E-commerce Q\&A}

\begin{table}[htbp]
\centering
\caption{Qualitative results of general e-commerce Q\&A}
\label{tab_qual_qa}
\begin{adjustbox}{width=1\linewidth}
\begin{tabular}{cc|p{7cm}}
\hline
\multicolumn{2}{c|}{Model}                                                           & Generated Answer        \\ \hline \hline
\multicolumn{2}{c|}{\color{olive}Instr.}                                             & \multicolumn{1}{l}{How to Use Etsy Stats for Your Shop?} \\ \hline
\multicolumn{2}{c|}{\color{orange}Answer}                                             & \multicolumn{1}{p{6.6cm}}{Etsy Stats provide you with data for your shop. With Stats, you can see how your shop is performing, where your traffic to your shop comes from and how shoppers are engaging with your listings.} \\ \hline \hline
\multicolumn{2}{c|}{GPT-3.5}                                                         & I can provide a list of ways to use Etsy Stats for your shop, as follows: 1. Track your shop's progress. 2. Learn about your customers. 3. Keep track of your promotions. 4. Analyze your item's performance. 5. Plan for the future.                     \\ \hline \hline
\multicolumn{2}{c|}{GPT-2}                                                           & Use Etsy Stats for your shop performance.                    \\ \hline
\multicolumn{2}{c|}{BART}                                                            & Use Etsy's Stats to track sales, categories, and shipments. Optimize and expand your campaign with data insights.                     \\ \hline \hline
\multicolumn{2}{c|}{T5-base}                                                         & To use Etsy Stats for Your Shop, you can follow these steps: 1. Choose a platform. 2. Create a website. 3. Set up your account on the Etsy platform. 4. Add content to your website. 5. Use analytics tools to track sales.                     \\ \hline
\multicolumn{2}{c|}{GPT-Neo}                                                          & You can use Etsy's latest features and features for your favorite handmade items, such as personalized designs, custom prints, and affiliate marketing                     \\ \hline
\multicolumn{1}{c|}{\multirow{11}{*}{\rotatebox{90}{\footnotesize{LLaMA}}}} & 7b  & Etsy Stats are important when it comes to running your shop. You can easily track the sales and view other statistics to help you know about your customers and business.                     \\ \cline{2-3} 
\multicolumn{1}{c|}{}                                                          & 13b & How often do you check your Etsy stats and shop page views in the last 30 days, or even 7 days time period? To maximize sales on Etsy, you should be checking these daily.                     \\ \cline{2-3} 
\multicolumn{1}{c|}{}                                                          & 30b & The guest post by Maria will share with us her tips and experience on how to use your stats on Etsy in order to help grow your business!                     \\ \hline \hline
\multicolumn{1}{c|}{\multirow{8}{*}{\rotatebox{90}{\footnotesize{LLaMA-E}}}} & \cellcolor{hiscolor}{7b}  & \cellcolor{hiscolor}{Etsy Stats: View sales, views, traffic, and search trends. Use to understand your shop's performance.}                     \\ \cline{2-3} 
\multicolumn{1}{c|}{}                                                          & 13b & Etsy Stats is used to calculate the shop performance like traffic sources, views, and favourites.                     \\ \cline{2-3} 
\multicolumn{1}{c|}{}                                                          & 30b & Etsy Stats tool is for analyzing your shop's performance and customer behaviour.                     \\ \hline 
\end{tabular}
\end{adjustbox}
\end{table}

This task can be regarded as a zero-shot evaluation since the testing questions are unseen in the training set. The results are in Table~\ref{tab_qual_qa}. Most general LLMs (such as T5-base, GPT-Neo, LLaMA-13/30b) are unable to effectively incorporate both the features of the e-commerce platform ("Etsy") and enquired entity ("Stats") in the given example, resulting in generated answers that are generic and semantically distant from the standard answers. In contrast, the LLaMA-E models generated answers all reflect the core semantics that this is a store evaluation tool provided by the platform, which is attributed to its background platform knowledge acquired through the related Q\&A instructions. We show more zero-shot generative held-out tasks of scenario-based ads generation and product recommendation in Appendix~\ref{sec_scag} and~\ref{sec_pur}.

The qualitative evaluation results indicate that content generated by LLaMA-E models effectively follows instructions and demonstrates the ability to infer complex semantics within the given scenario. For instance, the model completes the generative tasks while incorporating implicit semantic features on specific products by chain-of-thought reasoning that aligns with object conditions. These findings demonstrate that the object-interleaved instructions empower LLMs to associate semantics with relevant tasks while executing instructions, including recognizing common features shared by related products within the joint category. These abilities offer a thorough understanding of how LLaMA-E models are advantageous in e-commerce authoring scenarios encompassing correlated objects. Furthermore, the instructions enable LLMs to focus closely on the specific scenario features, preventing unnecessary generalizations to irrelevant tasks. This advantage facilitates providing professional and fine-grained object-oriented authoring services.

\section{Zero-shot Generalization Evaluation}
\label{sec_zero}

This section evaluates the model performance on unseen held-out tasks to test generalization in a zero-shot setting. This can be considered an evaluation of the models' applicability in real-world e-commerce authoring applications.

\subsection{Scenario-based Ads Generation}
\label{sec_scag}

This appendix presents examples of e-commerce authoring content generated by the LLaMA-E-7b model, demonstrating the best performance according to the $GM$ metric in the previous evaluation. The results are shown in Table \ref{tab_sag}. The evaluation aims to generate copywriting that aligns with the scenario features based on a set of products, which can be regarded as an assessment of the zero-shot performance on controllable text generation. We take the Christmas, Sports Fans, and Mid-year Sales as the scenarios of the festival, target customer group, and sales strategy for evaluation.

\begin{table}[htbp]
\centering
\caption{Examples of the generated scenario-adapted e-commerce authoring content. The understanding of the scenario features in responses is in bold font}
\label{tab_sag}
\begin{adjustbox}{width=1\linewidth}
\begin{tabular}{p{0.2cm}|p{0.2cm}|p{7.4cm}}
\hline
\multirow{10}{*}{\rotatebox{90}{\textbf{Christmas}}}     & \multirow{4}{*}{\rotatebox{90}{Inst.}} & Christmas is almost. Generate an ad for the following products: Himalayan salt lamp, bee house, grow sunflower plant kit, custom baby face mug, and puzzle plaque.                                                                                                                                                            \\ \cline{2-3} 
                               & \multirow{6}{*}{\rotatebox{90}{Resp.}}    & Enhance your home with a Himalayan salt lamp, bee house, grow sunflower plant kit, custom baby face mug, and puzzle plaque. Improve air quality, attract bees, and add a personal touch to your home with these unique \textbf{gifts for family and friends this holiday season}.                                                      \\ \hline \hline
\multirow{11}{*}{\rotatebox{90}{\textbf{Sports Fans}}}    & \multirow{4}{*}{\rotatebox{90}{Inst.}} & Generate an ad for sports fans based on the following products: football fact print, summer shorts, basketball youth shirts, swimming keychains, and energy chewing gum.                                                                                                                                                      \\ \cline{2-3} 
                               & \multirow{7}{*}{\rotatebox{90}{Resp.}}    & \textbf{Share your love for sports with our unique sports-themed gifts}. Order now and show your support for your \textbf{favourite athletes and teams} with our football fact print, summer shorts, basketball youth shirts, swimming keychains, and energy-boosting chewing gum. Elevate your game with our \textbf{sports-inspired gifts}. Order now! \\ \hline \hline
\multirow{8}{*}{\rotatebox{90}{\textbf{Mid-year Sale}}} & \multirow{4}{*}{\rotatebox{90}{Inst.}} & Generate a mid-year sale advertisement for the following products: hunting knife, green flame torch lighter, personalized wallet, and custom face socks.                                                                                                                                                                      \\ \cline{2-3} 
                               & \multirow{4}{*}{\rotatebox{90}{Resp.}}    & Elevate your \textbf{EDC} with a personalized wallet, hunting knife, green flame torch lighter, and custom face socks. \textbf{Shop now for a mid-year discount on your favourite gear}.                                                                                                                                                        \\ \hline
\end{tabular}
\end{adjustbox}
\end{table}

\subsection{Product Recommendation}
\label{sec_pur}

This appendix shows the capabilities of the LLaMA-E-7b model in the generative recommendation scenarios. The results are shown in Table \ref{tab_pur}. The model effectively summarizes the abstract semantic features within product collections through a comprehensive understanding of the products. It generates tailored recommendation content to meet specific features in customer instructions. These results reflect the model's diverse zero-shot recommendation performance, acquired through extensive exposure to vast commodity knowledge in the e-commerce domain. This LLMs-based method provides a novel recommendation paradigm with a broader range of flexible and diverse responses.

\begin{table}[htbp]
\centering
\caption{Examples of the generated e-commerce authoring content for product understanding and recommendation}
\label{tab_pur}
\begin{adjustbox}{width=1\linewidth}
\begin{tabular}{p{0.7cm}|p{6.8cm}}
\hline
Inst. & What products should I buy when planning my wedding?                                                                                                                              \\ \hline
Resp. & Etsy Wedding Supplies: Wedding Dress, Bridesmaid Dress, Flower Girl Dress, Invitation, Cake Topper, Candy Bar, Hair Accessories, Jewelry, Rings, Tiaras, Tuxedo, Groomsmen Gifts. \\ \hline \hline
Inst. & I'm looking for a gift for a middle-aged woman. Give me some ideas.                                                                                                               \\ \hline
Resp. & Etsy's Handmade, Vintage, and Unique Gifts for Her: Custom Jewelry, Pendants, and Bracelets, as well as Unusual Finds like Hair Accessories, Bags, and Rings for Her.             \\ \hline \hline
Inst. & I'm looking for a gift for a little kid. What should I choose between the "fire truck puzzle" and "slim cigarette case"?                                                          \\ \hline
Resp. & Fire truck puzzle for kids.                                                                                                                                                       \\ \hline \hline
Inst. & My friend recently bought a lot of things like leather cufflinks, bow ties, suits and brogue leather shoes. What do you think he is like?                                                                                                                                                                                  \\ \hline 
Resp. & He's a sophisticated gentleman who appreciates style and quality craftsmanship in clothing and accessories. He's a discerning buyer who knows what he's looking for and wants to make a statement with his purchases. He's a stylish man who wants to make a statement with his clothing and accessories.                                                                                                                                                                                  \\ \hline
\end{tabular}
\end{adjustbox}
\end{table}

\end{document}